\documentclass[aoas,MSNbibl,nameyear,seceqn,dvips]{arximspdf}
\usepackage{dcolumn}
\usepackage{graphicx}

\doi{10.1214/12-AOAS596} 
\volume{7}
\issue{1}
\pubyear{2013}
\firstpage{495}
\lastpage{522}

\makeatletter

\newcolumntype{d}[1]{D{.}{.}{#1}}

\newproclaim{algorithm}{Algorithm}[section]

\makeatother

\begin{document}
\begin{frontmatter}

\title{Regression trees for longitudinal and multiresponse data}
\runtitle{Regression trees}

\begin{aug}
\author[A]{\fnms{Wei-Yin} \snm{Loh}\corref{}\thanksref{t1}\ead[label=e1]{loh@stat.wisc.edu}\ead[label=u1,url]{http://www.stat.wisc.edu/\textasciitilde loh}}
\and
\author[B]{\fnms{Wei} \snm{Zheng}\ead[label=e2]{Wei.Zheng2@sanofi-aventis.com}}
\runauthor{W.-Y. Loh and W. Zheng}
\affiliation{University of Wisconsin--Madison and Sanofi-Aventis U.S. LLC}
\address[A]{Department of Statistics\\
University of Wisconsin--Madison\\
1300 University Avenue\\
Madison, Wisconsin 53706\\
USA\\
\printead{e1}\\
\printead{u1}}
\address[B]{Sanofi-Aventis U.S. LLC\\
300 3rd Street\\
Cambridge, Massachusetts 02139\\
USA\\
\printead{e2}} 
\end{aug}

\thankstext{t1}{Supported in part by U.S. Army Research
Office Grant W911NF-09-1-0205 and National Institutes of Health
Grant P50CA143188.}

\received{\smonth{8} \syear{2011}}
\revised{\smonth{8} \syear{2012}}

%
\begin{abstract}
Previous algorithms for constructing regression tree models for
longitudinal and multiresponse data have mostly followed the CART
approach. Consequently, they inherit the same selection biases and
computational difficulties as CART. We propose an alternative, based
on the GUIDE approach, that treats each longitudinal data series as
a curve and uses chi-squared tests of the residual curve patterns to
select a variable to split each node of the tree. Besides being
unbiased, the method is applicable to data with fixed and random
time points and with missing values in the response or predictor
variables. Simulation results comparing its mean squared prediction
error with that of MVPART are given, as well as examples comparing
it with standard linear mixed effects and generalized estimating
equation models. Conditions for asymptotic consistency of
regression tree function estimates are also given.
\end{abstract}

%

\end{frontmatter}

\section{Introduction}
\label{secintro}

A regression tree model is a nonparametric estimate of a regression
function constructed by recursively partitioning a data set with the
values of its predictor $X$ variables. CART [\citet{cart}] is one of
the oldest algorithms. It yields a piecewise-constant estimate by
recursively partitioning the data using binary splits of the form $X
\leq c$ if $X$ is ordinal, and $X \in A$ if $X$ is categorical. The
impurity of a node $t$ of the tree is defined as the sum of squared
deviations $i(t) = \sum(y-\bar{y}_t)^2$, where $\bar{y}_t$ is the
sample mean of response variable $Y$ in $t$ and the sum is over the
$y$ values in $t$. The split of $t$ into subnodes $t_L$ and $t_R$
that maximizes the reduction in node impurity $i(t) - i(t_L) - i(t_R)$
is selected. Partitioning continues until either the $X$ or the $y$
values are constant in a node, or the node sample size is below a
pre-specified threshold. Then the tree is pruned with the help of an
independent test sample or by cross-validation and the subtree with
the lowest estimated mean squared error is selected.

Several attempts have been made to extend CART to longitudinal and
multiresponse data, often by using likelihood-based functions as node
impurity measures. The earliest attempt for longitudinal data seems to
be \citet{Segal92jasa}, which uses the likelihood of an autoregressive
or compound symmetry model. If values are missing from the $Y$
variable, parameter estimation is performed by the EM algorithm.
Computational difficulties in estimating the covariance matrices limit
the method to data observed at equally-spaced time points.
\citet{ALSH02} follow the same approach, but use a likelihood-ratio
test statistic as the impurity function.

\citet{Zhang98} extends the CART approach to multiple binary response
variables, assuming there are no missing values in the $Y$
variable. It uses as impurity function the log-likelihood of an
exponential family distribution that depends only on the linear terms
and the sum of second-order products of the responses. \citet{Zhang08}
extend this idea to ordinal responses by first transforming them to
binary-valued indicator functions. Again, the approach is hindered by
the computational difficulties of having to compute covariance
matrices at every node.

\citet{death02} avoids the covariance computations by following the
CART algorithm exactly except for two simple modifications: the sample
mean is replaced by the $d$-dimensional sample mean and the node
impurity is replaced by $i(t) = \sum_{k=1}^d i_k(t)$, where $i_k(t)$
is the sum of squared deviations about the mean of the $k$th response
variable in $t$. The algorithm is implemented in the R package \mbox{MVPART}
[\citet{mvpart}]. \citet{LS04} adopt the same approach, but use the
Mahalanobis distance as node impurity, with covariance matrix
estimated from the whole data set.

In a different direction, \citet{YL99} treat each longitudinal data
vector as a random function or trajectory. Instead of fitting a
longitudinal model to each node, they first reduce the dimensionality
of the whole data set by fitting each data trajectory with a low-order
spline curve. Then they use the estimated coefficients of the basis
functions as multivariate responses to fit a regression tree model,
with the mean coefficient vectors as predicted values and standardized
squared error as node impurity. They recover the predicted trajectory
in each node by reconstituting the spline function from the mean
coefficient vector. They mention as an alternative the use of
principal component analysis to reduce the data dimension and then
fitting a multivariate regression tree model to the largest principal
components.

A major weakness of CART is that its selection of variables for splits
is biased toward certain types of variables. Because a categorical $X$
with $m$ unique values allows $(2^{m-1}-1)$ splits of the data and an
ordinal $X$ with $n$ unique values allows $(n-1)$ splits, categorical
variables with many unique values\vadjust{\goodbreak} tend to have an advantage over
ordinal variables in being selected [\citet{lohshih97,shih04,strobl07}].
This weakness is inherited by all multivariate extensions of CART,
including MVPART [\citet{HS07}]. Further, because reductions in node
impurity from splitting are based on observations without missing
values, variables with fewer missing values are more likely to yield
larger reductions (and hence be selected for splitting) than those
with more missing values; see Section~\ref{secmissing} below.

GUIDE [\citet{guide02}] avoids selection bias by replacing CARTs
one-step method of simultaneously selecting the split variable $X$ and
split set with a two-step method that first selects $X$ and then finds
the split set for the selected $X$. This approach makes it
practicable for GUIDE to fit a nonconstant regression model in each
node.

The goal of this article is to extend GUIDE to multivariate and
longitudinal response variables. Section~\ref{secguide} briefly
reviews the GUIDE variable selection method for univariate response
variables. Section~\ref{secmulti} extends it to multivariate
responses and longitudinal data observed at fixed time points. The
procedure is illustrated with an application to a data set on the
strength and viscosity of concrete. Section~\ref{secbias} compares
the selection bias and prediction accuracy of our method with MVPART
in a simulation study. Section~\ref{secmissing} deals with the
problem of missing values, which can occur in the predictor as well as
the response variables. We propose a solution and apply it to some
data on the mental health of children that are analyzed in
\citet{FLW04} with a generalized estimating equation (GEE) approach.
Section~\ref{seclongitudinal} further extends our method to
longitudinal data with random time points. We illustrate it with an
example on the hourly wages of high school dropouts analyzed in
\citet{SW03} with linear mixed effect (LME) models.
Section~\ref{secgee} compares the prediction accuracy of our method
with that of GEE and LME models in a simulation setting.
Section~\ref{secparallel} applies the ideas to simultaneously
modeling two longitudinal series from a study on maternal stress and
child illness analyzed in \citet{DHLZ02} with GEE logistic
regression. Section~\ref{sectheory} gives conditions for asymptotic
consistency of the multivariate regression tree function estimates and
Section~\ref{secconclusion} concludes the article with some remarks.

\section{Univariate GUIDE algorithm}
\label{secguide}
The GUIDE algorithm for a univariate response variable $Y$ can fit a
linear model in each node using one of several loss functions. For our
purposes here, it suffices to review the algorithm for least-squares
piecewise-constant models. The key idea is to split a node with the
$X$ variable that shows the highest degree of clustering in the signed
residuals from a constant model fitted to the data in the node. If a
predictor variable $X$ has no effect on the true regression mean
function, a plot of the residuals versus $X$ should not exhibit
systematic patterns. But if the mean is a function of $X$, clustering
of the signed residuals is expected.

\begin{figure}

\includegraphics{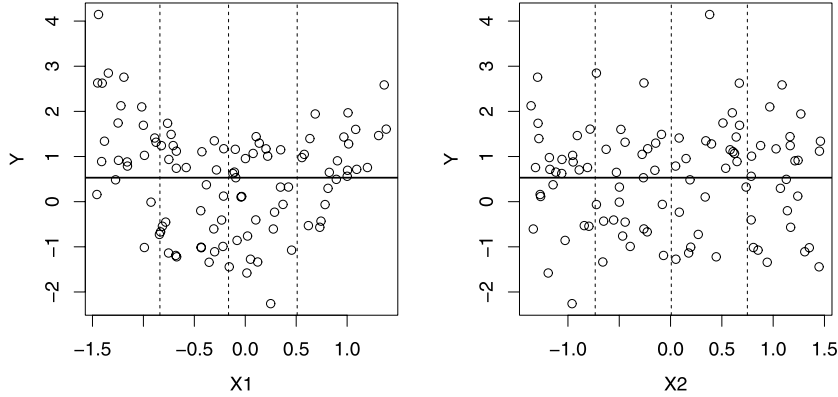}

\caption{Plots of $Y$ versus $X_1$ and $X_2$ from data generated
from the model $Y = X_1^2 + \varepsilon$. The horizontal line marks
the sample mean of $Y$.}
\label{figquad}
\end{figure}

To illustrate, consider some data generated from the model $Y = X^2_1
+ \varepsilon$, with $X_1$ and $X_2$ independent $U(-1.5, 1.5)$, that is,
uniformly distributed on the interval $(-1.5, 1.5)$, and $\varepsilon$
independent standard normal. Since the true regression function does
not depend on $X_2$, a piecewise-constant model should split on $X_1$
only. This is easily concluded from looking at plots of $Y$ versus
$X_1$ and $X_2$, as shown in Figure~\ref{figquad}. In the plot of $Y$
versus $X_1$, the positive residuals are clustered at both ends of the
range of $X_1$ and the negative residuals near the center. No such
clustering is obvious in the plot of $Y$ versus $X_2$.

\begin{table}[b]
\tablewidth=270pt
\caption{Contingency tables of $X_1$ and $X_2$ versus signs of residuals.
Chi-squared $p$-values are $0.005$ and $0.404$, respectively}
\label{tabquadcounts}
\begin{tabular*}{\tablewidth}{@{\extracolsep{\fill}}@{}lrrlrr@{}}
\hline
& \multicolumn{1}{c}{$\bolds{-}$} & \multicolumn{1}{c}{$\bolds{+}$}
& & \multicolumn{1}{c}{$\bolds{-}$} & \multicolumn{1}{c@{}}{$\bolds{+}$} \\
\hline
$-\infty< X_1 \leq-0.84$ & 5 & 17 & $-\infty< X_2 \leq-0.73$ & 9 & 16 \\
$-0.84 < X_1 \leq-0.16$ & 16 & 12 & $-0.73 < X_2 \leq0.01$ & 14 & 11 \\
$-0.16 < X_1 \leq0.51$ & 17 & 10 & $0.01 < X_2 \leq0.75$ & 9 & 16 \\
$0.51 < X_1 < \infty$ & 6 & 17 & $0.75 < X_2 < \infty$ & 12 & 13 \\
\hline
\end{tabular*}
\end{table}

GUIDE measures the degree of clustering by means of contingency table
chi-squared tests. In each test, the values of $X$ are grouped into a
small number of intervals (indicated by the vertical dashed lines in
Figure~\ref{figquad}), with the groups forming the rows and the
residual signs forming the columns of the table. The end points are
computed such that each interval has approximately the same number of
observations if $X$ is uniformly distributed (see
Algorithm~\ref{algmult} below for the definitions).
Table~\ref{tabquadcounts} shows the table counts and the chi-squared
$p$-values for the data in Figure~\ref{figquad}. If $X$ is a
categorical variable, its values are used to form the rows of the
table.

GUIDE selects the variable with the smallest chi-squared $p$-value to
split the node. Because the sample size in a node decreases with
splitting, the $p$-values are approximate at best. Their exact values
are not important, however, as they serve only to rank the variables
for split selection. Similar $p$-value methods have been used in
classification tree algorithms, for example, F-tests [\citet{lohshih97}] and
permutation tests [\citet{party}]. One benefit from using $p$-values is
lack of selection bias, at least for sufficiently large sample
sizes. This is due to the $p$-values being approximately identically
distributed if all the $X$ variables are independent of $Y$.

After a variable is selected, the split set is found by exhaustive
search to maximize the reduction in the sum of squared residuals. A side
(but practically important) benefit is significant computational
savings over the CART method of searching for the best split set for
\textit{every} $X$. The procedure is applied recursively to construct
an overly large tree. Then the tree is pruned using cross-validation
as in the CART algorithm and the subtree with the smallest
cross-validation estimate of mean squared error is selected.

\section{Multiple response variables}
\label{secmulti}

We motivate our extension of GUIDE to multiple response variables with
an analysis of some data on the strength and viscosity of concrete
[\citet{Yeh07}] taken from the UCI Machine Learning Repository
[\citet{AsuncionNewman2007}]. There are 103 complete observations on
seven predictor variables (cement, slag, fly ash, water,
superplasticizer (SP), coarse aggregate and fine aggregate, each
measured in kg per cubic meter) and three response variables (slump
and flow, in cm, and 28-day compressive strength in Mpa). Slag and fly
ash are cement substitutes. Slump and flow measure the viscosity of
concrete; slump is the vertical height by which a cone of wet concrete
sags and flow is the horizontal distance by which it spreads. The
objective is to understand how the predictor variables affect the
values of the three response variables jointly.

\begin{table}
\caption{Separate linear regression models,
with $p$-values less than 0.05 in italics}
\label{tabconcmlr}
\begin{tabular*}{\tablewidth}{@{\extracolsep{\fill}}ld{3.3}cd{4.3}cd{3.3}c@{}}
\hline
& \multicolumn{2}{c}{\textbf{Slump}} & \multicolumn{2}{c}{\textbf{Flow}} &
\multicolumn{2}{c@{}}{\textbf{Strength}} \\[-4pt]
& \multicolumn{2}{c}{\hrulefill} & \multicolumn{2}{l}{\hspace*{-2pt}\hrulefill} &
\multicolumn{2}{c@{}}{\hrulefill} \\
& \multicolumn{1}{c}{\textbf{Estimate}} & \multicolumn{1}{c}{$\bolds{p}$\textbf{-value}}
& \multicolumn{1}{c}{\textbf{Estimate}} &
\multicolumn{1}{c}{$\bolds{p}$\textbf{-value}} & \multicolumn{1}{c}{\textbf{Estimate}}
& \multicolumn{1}{c@{}}{$\bolds{p}$\textbf{-value}} \\
\hline
(Intercept)& -88.525 & 0.66 & -252.875 & 0.47 & 139.782 & 0.052\\
Cement & 0.010 & 0.88 & 0.054 & 0.63 & 0.061 & \textit{0.008}\\
Slag & -0.013 & 0.89 & -0.006 & 0.97 & -0.030 & 0.352\\
Fly ash & 0.006 & 0.93 & 0.061 & 0.59 & 0.051 & \textit{0.032}\\
Water & 0.259 & 0.21 & 0.732 & \textit{0.04} & -0.233 & \textit{0.002}\\
SP & -0.184 & 0.63 & 0.298 & 0.65 & 0.10 & 0.445\\
CoarseAggr & 0.030 & 0.71 & 0.074 & 0.59 & -0.056 & \textit{0.045}\\
FineAggr & 0.039 & 0.64 & 0.094 & 0.51 & -0.039 & 0.178\\ \hline
\end{tabular*}
\end{table}

\begin{figure}[b]

\includegraphics{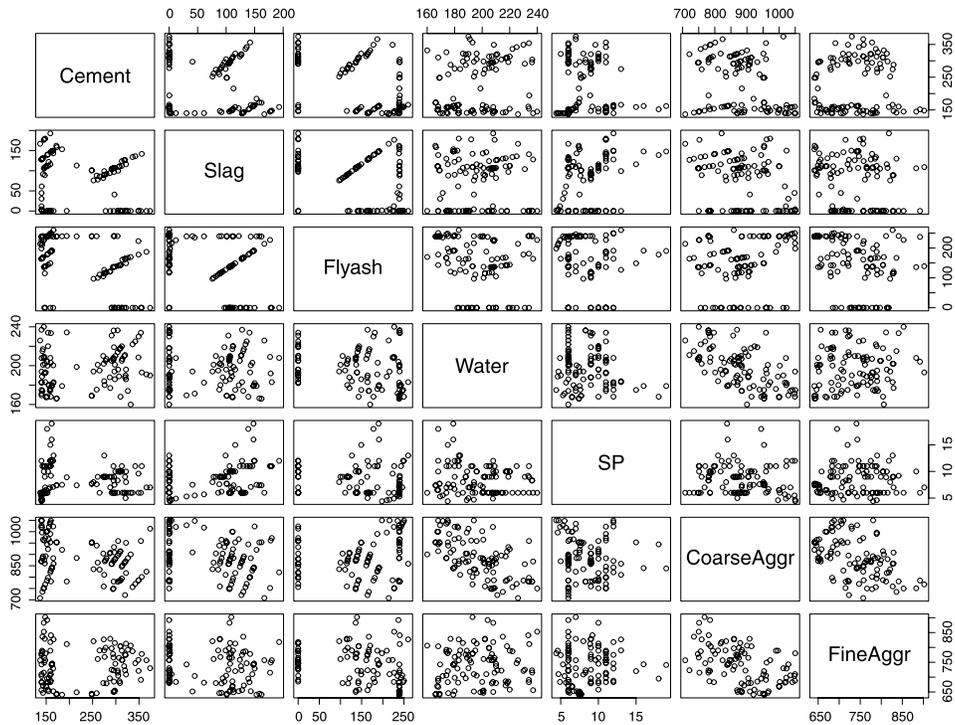}

\caption{Plots of pairs of predictor variables for concrete data.}
\label{figconcpairs}
\end{figure}

Fitting a separate multiple linear regression model to each response
is not enlightening, as the results in Table~\ref{tabconcmlr} show.
Cement, fly ash, water and coarse aggregate are all significant (at
the 0.05 level) for strength. The signs of their coefficients suggest
that strength is increased by increasing the amounts of cement and fly
ash and decreasing that of water and coarse aggregate. Since no
variable is significant for slump, one may be further tempted to
conclude that none is important for its prediction. This is false,
because a linear regression for slump with only water and slag as
predictors finds both to be highly significant. The problem is due to
the design matrix being quite far from orthogonal (see
Figure~\ref{figconcpairs}). Therefore, it is risky to interpret each\vadjust{\goodbreak}
regression coefficient by ``holding the other variables constant.''
Besides, the main effect models are most likely inadequate anyway.
Inclusion of interaction terms, however, brings on other difficulties,
such as knowing which terms and of what order to add, which makes
interpretation even more challenging.

\begin{figure}

\includegraphics{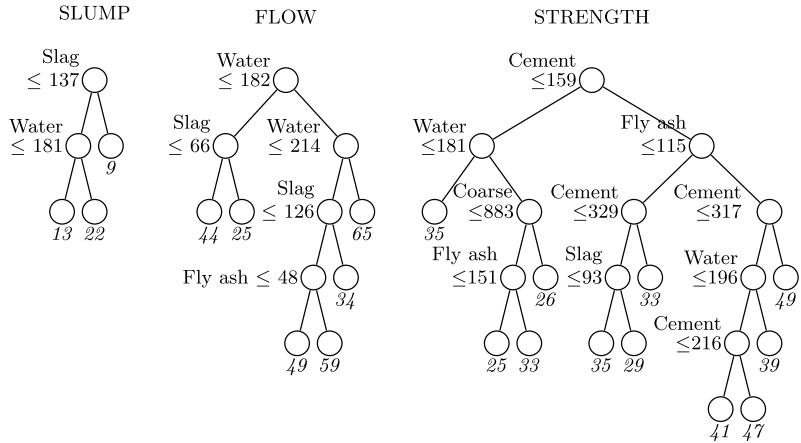}

\caption{Univariate GUIDE models for predicting slump (left), flow
(middle) and strength (right) of concrete. At each node, a case
goes to the left subnode if and only if the stated condition is
satisfied. The predicted value is in italics below each terminal
node.}
\label{figconcsingle}
\end{figure}

Instead of controlling for the effects of other variables by means of
an equation, a regression tree model achieves a similar goal by
dividing the sample space into partitions defined by the values of the
variables, thus effectively restricting the ranges of their values.
Figure~\ref{figconcsingle} shows three GUIDE tree models, one for
each response variable, with predicted values beneath the terminal
nodes. We see that less slag and more water yield larger values of
slump, more water yields larger values of flow, and higher amounts of
cement and fly ash produce the strongest concrete. Although it is
easier to interpret the tree structures than the coefficients of the
linear models, it is still nontrivial to figure out from the three
trees how the variables affect the response variables jointly. For
example, the trees show that (i) slump is least when slag${}>{}$137,
(ii) flow is least when water${}\leq{}$182 and slag${}>{}$66, and
(iii) strength is greatest when cement${}>{}$317 and fly ash${}>{}$115.
We may thus conclude that the intersection of these conditions yields
the strongest and least viscous concrete. But there are no
observations in the intersection.

A single tree model that simultaneously predicts all three responses
would not have these difficulties. Ideally, such an algorithm would
produce compact trees with high predictive accuracy and without
variable selection bias. The main hurdle in extending GUIDE to
multiple response variables is unbiased variable selection. Once this
problem is solved, the rest of the method follows with a simple
modification to the node impurity function.

\citet{Lee05} proposes one extension, applicable to ordinal $X$
variables only, that fits a GEE model to the data in each node. It
classifies each observation into one of two groups according to the
sign of its average residual, $d^{-1}\sum_{k=1}^d
\hat{\varepsilon}_{ik}$, where $\hat{\varepsilon}_{ik}$ is the residual of
the $i$th observation for the $k$th response variable. Then a
two-sample $t$-test is performed for each $X$ and the one with the
smallest $p$-value is selected to split the node. The split point is a
weighted average of the $X$ values in the two groups. If the smallest
$p$-value exceeds a pre-specified threshold, splitting stops.

Lee's solution is deficient in several respects. First, the $p$-value
threshold is hard to specify, because it depends on characteristics of
the data set, such as the number and type of variables and the sample
size. Second, it is inapplicable to categorical predictor
variables. Third, it is inapplicable to data with missing predictor or
response values. Finally, for the ultimate goal of clustering the
response vectors into groups with similar patterns, classifying them
into two groups by the signs of their \textit{average} residuals is
potentially ineffective, because two response vectors can have very
dissimilar patterns and yet have average residuals with the same sign.

A more effective extension can be obtained by working with the
residual sign vectors instead. Let $(Y_1, Y_2,\ldots, Y_d)$ be the
$d$ response variables. At each node, we fit the data with the sample
mean vector and compute the residual vectors. Since each residual can
have a positive or nonpositive sign, there are $2^d$ possible
patterns for the residual sign vector. To determine if a predictor
variable $X$ is independent of the residual pattern, we form a
contingency table with the sign patterns as the columns and the
(grouped, if $X$ is not categorical) values of $X$ as the rows and
find the $p$-value of the chi-squared test of independence. Specific
details are given in the algorithm below. Other aspects of the method
are the same as in the univariate GUIDE, except for the node impurity
function being the sum of (normalized, if desired) squared
errors.\vspace*{-3pt}
%
\begin{algorithm} \label{algmult} Split variable selection at each
node $t$:
\renewcommand\theenumi{(\arabic{enumi})}
\renewcommand\labelenumi{\theenumi}
\begin{enumerate}
\item Find $\bar{y} = (\bar{y}_1,\ldots, \bar{y}_d)$, where
$\bar{y}_k$ is the mean of the nonmissing values of the $k$th
response variable in $t$.
\item\label{itemmissy} Define the sign vector $Z = (Z_{1},Z_{2},\ldots, Z_{d})$ such
that $Z_{k} = 1$ if $Y_k > \bar{y}_k$ and $Z_{k} = -1$ if $Y_k
\leq\bar{y}_k$. If $Y_k$ is missing, the user can choose either
$Z_k = 1$ or $Z_k = -1$ (default), with the same choice used for all
nodes.
\item Main effect tests. Do this for each $X$ variable:
\begin{enumerate}[(a)]
\item[(a)]
\hypertarget{stepmain1}
If $X$ is not categorical, group its
values into $m$ intervals. Let $\bar{x}$ and $s$ denote the
sample mean and standard deviation of the nonmissing values of
$X$ in~$t$. If the number of data points is less than $5 \times
2^{d+2}$, set $m=3$ and define the interval end points to be
$\bar{x} \pm s\sqrt{3}/3$. Otherwise, set $m=4$ and define the
interval end points as $\{\bar{x}, \bar{x} \pm s\sqrt{3}/2\}$.
\item[(b)] If $X$ is categorical, use its categories to form the
groups.
\item[(c)]
\hypertarget{itemmissx} Create an additional group for missing values if $X$ has
any.
\item[(d)] Form a contingency table with the $2^d$ patterns of $Z$ as
columns and the $X$-groups as rows, and compute the $p$-value of
the chi-squared test of independence.
\end{enumerate}
\item\label{stepmain} If the smallest $p$-value is less than $0.05/d$, select the
associated $X$ variable and exit.\vadjust{\goodbreak}
\item\label{stepinteract} Otherwise, do these interaction tests
for each pair of variables $X_i, X_j$:
\begin{enumerate}[(a)]
\item[(a)] If $X_i$ is noncategorical, split its range into two
intervals $A_{i1}$ and $A_{i2}$ at its sample mean. If $X_i$ is
categorical, let $A_{ik}$ denote the singleton set containing
its $k$th value. Do the same for $X_j$.
\item[(b)] Define the sets $B_{k,m}\,{=}\,\{(x_i, x_j)\dvtx  x_i\!\in\!
A_{ik}, x_j\!\in\!A_{jm}\}$, for $k, m\,{=}\,1,2, \ldots\,$.
\item[(c)] Form a contingency table with the $Z$ patterns as columns
and $\{B_{k,m}\}$ as rows and compute its $p$-value.
\end{enumerate}
\item\label{stepinteractexit}
If the smallest $p$-value from the interaction tests is less
than $0.05/\{d(d-1)\}$, select the associated pair of predictors.
\item Otherwise, select the $X$ with the smallest main effect
$p$-value from step~\ref{stepmain}.
\end{enumerate}
\end{algorithm}
The value of $m$ in step \hyperlink{stepmain1}{(3)(a)} is chosen to keep the
row-dimension of the table as small as possible without sacrificing
its ability to detect patterns. The interval end points are chosen so
that if $X$ has a uniform distribution, each interval has roughly the
same number of observations. If $d=1$, these definitions reduce to
those in the univariate GUIDE algorithm [\citet{guide8}, page 1716].

If a noncategorical variable is selected in step~\ref{stepmain}, the
split $X \leq c$ is found by searching over all midpoints $c$ of
consecutive order statistics to minimize the total sum of squared
deviations of the the two subnodes. If $X$ is a categorical variable,
the search for a split of the form $X \in A$ can be computationally
daunting if $X$ takes many values. To obtain a quick but approximate
solution, we create a classification variable from the $Z$ patterns in
each node and then use the method described in \citeauthor{guide8}
[(\citeyear{guide8}), the Appendix] for classification trees to find the
set $A$. We also use the procedures in \citet{guide8} to find the
split set if a pair of variables is selected in step
\ref{stepinteractexit}.\looseness=-1

\begin{table}[b]
\tablewidth=310pt
\caption{Contingency table formed by cross-tabulating residual
signs vs. water groups}
\label{tabconccontab}
\begin{tabular*}{\tablewidth}{@{\extracolsep{\fill}}rcccccccc@{}}
\hline
$Z_1$ & $-$ & $-$ & $-$ & $-$ & $+$ & $+$ & $+$ & $+$ \\
$Z_2$ & $-$ & $-$ & $+$ & $+$ & $-$ & $-$ & $+$ & $+$ \\
$Z_3$ & $-$ & $+$ & $-$ & $+$ & $-$ & $+$ & $-$ & $+$ \\ \hline
\multicolumn{1}{@{}l}{Water${}\leq{}$185.5} & 5 & 16 & 0 & 0 & 1 & 4 & \hphantom{0}6 & \hphantom{0}2\\
185.5${}<{}$Water${}\leq{}$208.8 & 6 & \hphantom{0}2 & 1 & 0 & 4 & 1 & 14 & 13\\
\multicolumn{1}{@{}l}{Water${}>{}$208.8} & 3 & \hphantom{0}2 & 1 & 1 & 0 & 0 & 13 & \hphantom{0}8\\
\hline
\end{tabular*}
\end{table}

%
\begin{figure}

\includegraphics{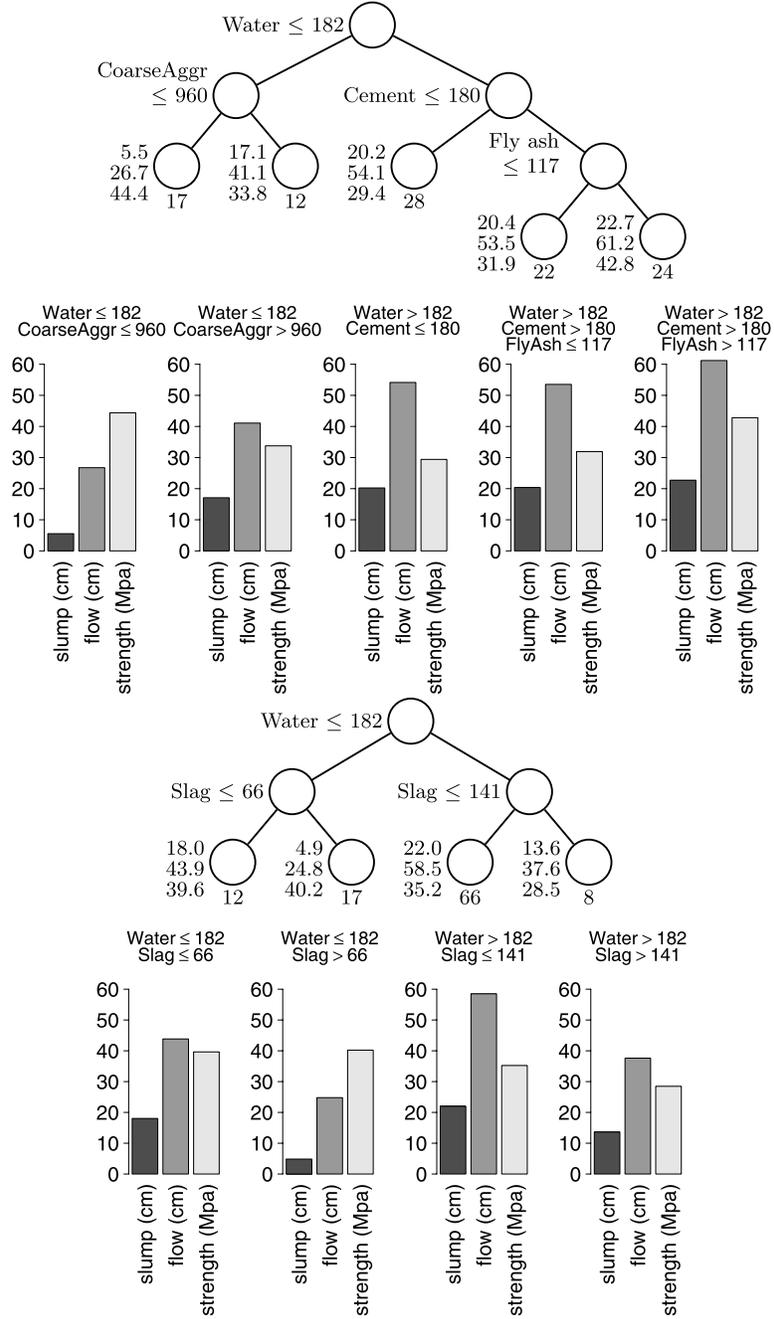}

\caption{Multivariate GUIDE (top) and MVPART (bottom) models for
the concrete data. Sample sizes are beneath and predicted values
(slump, flow and strength, resp.) are on the left of each node.
Barplots show the predicted values in the terminal nodes of the
trees.}
\label{figconctrees}
\end{figure}

For the concrete data, the values of each $X$ variable are grouped
into three intervals. Table~\ref{tabconccontab} shows the
contingency table formed by the residual signs and the groups for
water, which has the smallest chi-squared $p$-value of $8 \times
10^{-5}$. The top half of Figure~\ref{figconctrees} shows the tree
model after pruning by ten-fold cross-validation. We will use the
description ``multivariate GUIDE'' to\vadjust{\goodbreak} refer to this method from now
on. Predicted values of the response variables are shown by the
heights of the bars in the figure. The strongest and most viscous
concrete is obtained with water${}\leq{}$182~kg/m$^3$ and coarse
aggregate${}\leq{}$960 kg/m$^3$. This is consistent with these two
variables having negative coefficients for strength in
Table~\ref{tabconcmlr}. The tree model also shows that the
combination of water${}>{}$182 kg/m$^3$, cement${}>{}$180~kg/m$^3$ and
fly ash${}>{}$117 kg/m$^3$ yields concrete that is almost as strong but
least viscous. Thus, it is possible to make strong concrete with low or
high viscosity. The combination predicting concrete with the least
strength is water${}>{}$182 kg/m$^3$ and cement${}\leq{}$180~kg/m$^3$.
The MVPART [\citet{mvpart}] model is shown in the bottom half of
Figure~\ref{figconctrees}. Its first split is the same as that of
GUIDE, but the next two splits are on slag.

To compare the prediction accuracy of the methods, we first normalize
the values of the three response variables to have zero mean and unit
variance and then apply leave-one-out cross-validation to estimate
their sum of mean squared prediction errors of the pruned trees, where
the sum is over the three response variables. The results are quite
close, being 1.957, 2.097 and 2.096 for univariate GUIDE,
multivariate GUIDE and MVPART, respectively. As we will see in the
next section, univariate trees tend to have lower prediction error
than multivariate trees if the response variables are uncorrelated and
higher prediction error when the latter are correlated. In this
example, slump and flow are highly correlated (cor${}={}$0.91) but each is
weakly correlated with strength ($-0.22$ and $-0.12$, resp.). Thus,
there is a cancellation effect.

\section{Selection bias and prediction accuracy}
\label{secbias}
\begin{figure}

\includegraphics{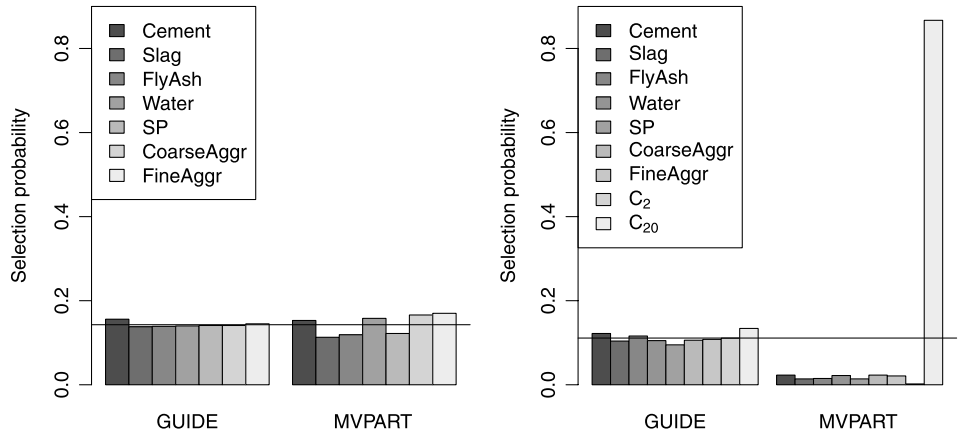}

\caption{Estimated probabilities (based on 5000 simulation trials)
that each predictor variable is selected to split the root node
when all are independent of the response variables. Standard
errors are less than 0.005. Variable $C_k$ is multinomial with
equal probabilities on $k$ categories. The horizontal line marks
the level for unbiased selection.}
\label{figconcbias}
\end{figure}

We carried out some simulation experiments to further compare the
variable selection bias and prediction accuracy of GUIDE and MVPART.
To show the selection bias of MVPART, we took the concrete data as a
population distribution and drew bootstrap samples from it of the same
size ($n = 103$). Then we randomly permuted the values in each
predictor variable to render it independent of the response
variables. An unbiased algorithm now should select each variable with
the same probability $1/7 = 0.143$ to split the root node. The left
panel of Figure~\ref{figconcbias} shows the estimated selection
probabilities for GUIDE and MVPART from 5000 simulation trials. The
estimates for GUIDE are all within two simulation standard errors of
$1/7$ but those of MVPART are not: they are roughly proportional to the
number of unique values of the variables in the data, namely, 80, 63,
58, 70, 32, 92 and 90 for cement, slag, fly ash, water, SP, coarse
aggregate and fine aggregate, respectively.

To demonstrate the bias of MVPART toward selecting variables with more
split sets, we added two independent predictor variables, $C_2$ and
$C_{20}$, where $C_k$ denotes a multinomial variable with equal
probabilities on $k$ categories. Variable $C_2$ allows only one split
but variable $C_{20}$ has $2^{19}-1 = 524$,287
splits. An unbiased method now should select each
variable with probability $1/9 = 0.111$. The results, based on 5000
simulation trials, are shown in the right panel of Figure
\ref{figconcbias}. GUIDE is again essentially unbiased (within
simulation error), but MVPART selects $C_{20}$ more than 86\% of the
time and $C_2$ only 10 out of 5000 times.

To compare the prediction accuracies of MVPART and univariate and
multivariate GUIDE, we use three simulation scenarios, with each
having seven predictor variables and three response variables. The
values of the response variables are generated by the equation $Y_k =
\mu_k + \varepsilon$, $k = 1, 2, 3$, where the $\varepsilon$ are independent
normal variables with mean 0 and variance 0.25. The three scenarios
are as follows:
%
\begin{eqnarray}
\label{eqsim4}
(\mu_1, \mu_2, \mu_3) & = &
(X_1, X_2, X_3),
\\
\label{eqsim5}
(\mu_1, \mu_2, \mu_3) & = &
(X_1 + X_2, X_1 + X_2,
X_1 + X_2),
\\
\label{eqsim2}
(\mu_1, \mu_2, \mu_3) & = & \cases{ (1, -1,
0), &\quad $X_1X_2 > 0$,
\cr
(0, 0, 1), &\quad
$X_1X_2 \leq0$. }
\end{eqnarray}
Scenarios (\ref{eqsim4}) and (\ref{eqsim5}) are standard linear
regression models. Univariate GUIDE should be most accurate in
scenario (\ref{eqsim4}), because each mean response depends on a
different predictor variable. The same may not be true for
scenario (\ref{eqsim5}), where a multivariate regression tree may be
able to utilize the joint information among the response variables.
Scenario (\ref{eqsim2}) has a piecewise-constant tree structure, but
it can be challenging due to the absence of main effects.

Two simulation experiments were performed. In the first experiment,
variables $X_1,\ldots, X_7$ are mutually independent $U(-0.5, 0.5)$.
For each scenario, 100 training samples are generated in each
simulation trial and a pruned regression tree model (using the CART
pruning method) constructed by each method. One hundred independent
test values $(X_{1j}, X_{2j},\ldots, X_{7j})$, $j=1,\ldots,100$, are
generated to evaluate the models. Let $(\mu_{1j}, \mu_{2j},
\mu_{3j})$ denote the mean response values for the $j$th test sample,
$(\hat{\mu}_{1j}, \hat{\mu}_{2j}, \hat{\mu}_{3j})$ denote their
predicted values, and $\mathrm{MSE} = \sum_{j=1}^{100} \sum_{i=1}^3
(\hat{\mu}_{ij}-\mu_{ij})^2/100$ denote the estimated mean squared
error.

\begin{table}
\caption{Estimated mean squared error (MSE) and number
of terminal nodes (Nodes) using 100 training samples in 1000 simulation trials.
Standard errors of MSE in parentheses.
``Univariate GUIDE'' refers to the model with a separate tree for each
response variable}
\label{tabsimresults}
\begin{tabular*}{\tablewidth}{@{\extracolsep{\fill}}lcd{2.1}cccd{2.1}@{}}
\hline
& \multicolumn{2}{c}{\textbf{Univariate GUIDE}} &
\multicolumn{2}{c}{\textbf{Multivariate GUIDE}}
& \multicolumn{2}{c@{}}{\textbf{MVPART}} \\[-4pt]
& \multicolumn{2}{c}{\hrulefill} & \multicolumn{2}{c}{\hrulefill}
& \multicolumn{2}{c@{}}{\hrulefill} \\
\textbf{Scenario} & \multicolumn{1}{c}{$\bolds{\mathrm{MSE}\times10^2}$} &
\multicolumn{1}{c}{\textbf{Nodes}} & \multicolumn{1}{c}{$\bolds{\mathrm{MSE} \times
10^2}$} & \multicolumn{1}{c}{\textbf{Nodes}}
& \multicolumn{1}{c}{$\bolds{\mathrm{MSE} \times10^2}$}
& \multicolumn{1}{c@{}}{\textbf{Nodes}} \\
\hline
& \multicolumn{6}{c@{}}{$X_1,\ldots, X_7$ are independent $U(-0.5,
0.5)$} \\
[4pt]
(\ref{eqsim4}) & \hphantom{0}14.1 (0.1) & 5.7 & \hphantom{0}21.9 (0.1) & 3.4 & \hphantom{0}22.2 (0.1) &
3.1 \\
(\ref{eqsim5}) & \hphantom{0}35.1 (0.2) & 8.3 & \hphantom{0}24.2 (0.2) & 4.5 & \hphantom{0}22.3 (0.2) &
4.5 \\
(\ref{eqsim2}) & \hphantom{0}33.0 (0.5) & 11.8 & \hphantom{0}12.3 (0.4) & 4.2 & \hphantom{0}68.8 (0.7) &
2.7  \\
[4pt]
& \multicolumn{6}{c@{}}{$X_1,\ldots, X_6$ are $N(0, V)$, $X_7$ is independent
$U(-0.5, 0.5)$}  \\
[4pt]
(\ref{eqsim4}) & \hphantom{0}47.2 (0.3) & 13.7 & 156.0 (0.7) & 6.4 & 128.4 (0.6)
& 13.0 \\
(\ref{eqsim5}) & 198.0 (1.2) & 22.5 & 206.1 (1.4) & 6.3 & 158.2
(1.2) & 10.9 \\
(\ref{eqsim2}) & \hphantom{0}48.0 (0.6) & 11.3 & \hphantom{0}15.5 (0.5) & 4.4 & \hphantom{0}68.2 (0.8) &
3.6 \\
\hline
\end{tabular*}
\end{table}

The upper half of Table~\ref{tabsimresults} shows the average values
of MSE and their standard errors over 1000 simulation trials. The
average numbers of terminal nodes are also shown (for the univariate
GUIDE method, this is the sum of the number of terminal nodes of the
separate trees). As expected, univariate GUIDE is more accurate than
the multivariate tree methods in scenario (\ref{eqsim4}), where the
means are unrelated. On the other hand, multivariate GUIDE is more
accurate in scenarios (\ref{eqsim5}) and (\ref{eqsim2}) because it
can take advantage of the relationships among the response variables.
The accuracy of MVPART is close to that of multivariate GUIDE, except
in scenario (\ref{eqsim2}), where it has difficulty detecting the
interaction effect. The higher accuracy of multivariate GUIDE here is
due to the interaction tests in step~\ref{stepinteract} of
Algorithm~\ref{algmult}.

In the second experiment, we generated $(X_1,\ldots, X_6)$ as
multivariate normal vectors with zero mean and covariance matrix
\[
V = %
\pmatrix{ 1 & 0 & r & r & 0 & 0
\cr
0 & 1 & 0 & 0 & r & r
\cr
r &
0 & 1 & r & 0 & 0
\cr
r & 0 & r & 1 & 0 & 0
\cr
0 & r & 0 & 0 & 1 & r
\cr
0 & r &
0 & 0 & r & 1 } \vadjust{\goodbreak}%
\]
and $r= 0.5$. Thus, $(X_1, X_3, X_4)$ is independent
of $(X_2, X_5, X_6)$. As in the previous experiment, $X_7$ is
independent $U(-0.5, 0.5)$. The results, given in the bottom half of
the table, are quite similar to those in the first experiment, except
in scenario (\ref{eqsim4}), where MVPART has lower MSE than
multivariate GUIDE, and in scenario (\ref{eqsim5}), where univariate
GUIDE has lower MSE than multivariate GUIDE. Notably, the average
number of terminal nodes in the MVPART trees is about twice the
average for multivariate GUIDE in these two scenarios. The larger
number of nodes suggest that the trees may be splitting on the wrong
variables more often. But because these variables are correlated with
the correct ones and because of the effectiveness of pruning, the MSEs
are not greatly increased.

\section{Missing values}
\label{secmissing}
Missing values in the predictor variables do not present new
challenges, as the method in univariate GUIDE can be used as follows
[\citet{guide8}]. If $X$ has missing values, we create a ``missing''
group for it and carry out the chi-squared test with this additional
group. Besides allowing all the data to be used, this technique can
detect relationships between the missing patterns of $X$ and the
values of the response variables.

The search for a split set for a categorical $X$ with missing values
is no different from that for a categorical variable without missing
values, because missing values are treated as an additional category.
But if $X$ is noncategorical and has missing values, we need to find
the split point and a method to send cases with missing values through
the split. For the first task, all splits at midpoints between
consecutive order statistics of $X$ are considered. All missing $X$
values are temporarily imputed with the mean of the nonmissing values
in the node. Because the sample mean usually belongs to the node with
the greater number of observations, this typically sends the missing
values to the larger node. The best among these splits is then
compared with the special one that sends all missing values to one
node and all nonmissing values to the other, and the one yielding the
greater impurity reduction is selected.

\begin{figure}

\includegraphics{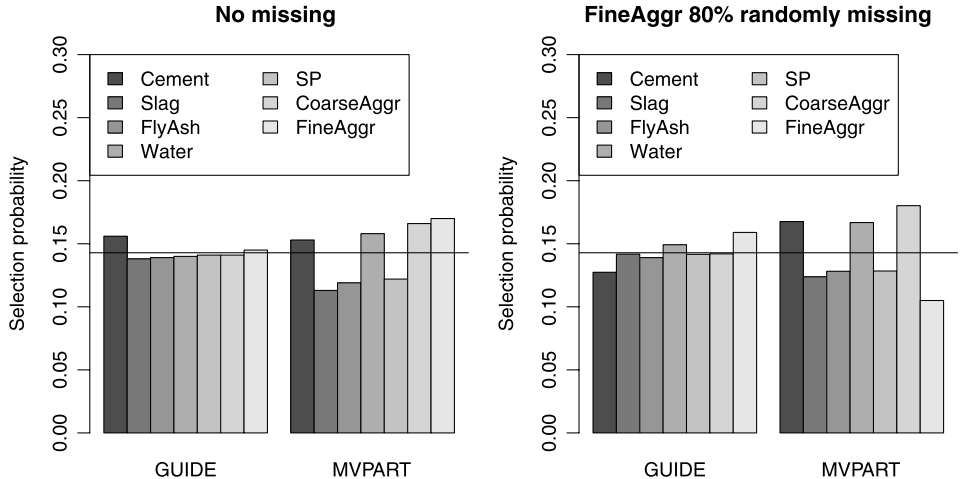}

\caption{Estimated probabilities (from 5000 simulation trials) of
variable selection when all variables are independent of the
response variables. Standard errors are less than 0.005. The
horizontal line marks the probability for unbiased selection.}
\label{figconcbiasmiss}
\end{figure}

Our approach to split selection is different from that of MVPART,
which uses the CART method of searching for the split that maximizes
the reduction in total sum of squared errors \textit{among the
observations nonmissing in the split variable}. As a consequence,
MVPART has a selection bias toward variables with \textit{fewer} missing
values. This can be demonstrated using the procedure in
Section~\ref{secbias}, where we take bootstrap samples of the
concrete data and randomly permute its predictor values.
Figure~\ref{figconcbiasmiss} shows the selection probabilities
before and after 80\% of the values in FineAggr are made randomly
missing, based on 5000 simulation trials. Variations in the GUIDE
probabilities are all within three simulation standard errors of $1/7$,
but those of MVPART are not. More importantly, there is a sharp drop
in the selection probability of FineAggr due to missing values.

Missing values in a univariate response variable do not cause
problems, because those observations are routinely omitted. But if
the response is multidimensional, it is wasteful to omit an
observation simply because one or more responses are missing, as
MVPART and the methods of \citet{ALSH02} and \citet
{Zhang98} require.
\citet{Segal92jasa} allows missing responses in longitudinal
data, but
only if the variable is continuous, is observed at equally-spaced time
points, and the data in each node are fitted with an autoregressive or
compound symmetry model. In our approach, if there are missing values
in some but not all response variables, step~\ref{itemmissy} of
Algorithm~\ref{algmult} takes care of them by giving the user the
choice of $Z_k = -1$ or $Z_k = 1$ for missing $Y_k$. For split set
selection, we compute the mean response for each $Y_k$ from the
nonmissing values in the node and the sum of squared errors from the
nonmissing $Y_k$ values only.

To illustrate these ideas, consider a data set from a survey of the
mental health of 2501 children, analyzed in
\citet{FLW04}, Section 16.5. One purpose of the survey was to understand
the influence of parent status (single vs. not single) and child's
physical health (good vs. fair or poor) on the prevalence of
externalizing behavior in the child. Each child was assessed
separately by two ``informants'' (a parent and a teacher) on the
presence or absence (coded 1 and 0, resp.) of delinquent or
aggressive externalizing behavior. All the parent responses were
complete, but 1073 children (43\%) did not have teacher responses.

For child $i$, let $Y_{ij} = 1$ if the $j$th informant (where $j=1$
refers to parent and $j=2$ to teacher) reports externalizing behavior,
and $Y_{ij} = 0$ otherwise. Assuming that the $Y_{ij}$ are missing at
random and the covariance between the two responses is constant,
Fitzmaurice et al. use a generalized estimating equation (GEE) method to
simultaneously fit this logistic regression model to the two
responses:
\[
\log\bigl\{P(Y_{ij}=1) / P(Y_{ij}=0)\bigr\} =
\beta_0 + \beta_1 x_{1ij} + \beta_2
x_{2ij} + \beta_3 x_{3ij} + \beta_{13}
x_{1ij} x_{3ij}.
\]
Here $x_{1ij} = 1$ if $j=1$ and 0 otherwise, $x_{2ij} = 1$ if the
parent is single and 0 otherwise, and $x_{3ij} = 1$ if the child's
health is fair or poor and 0 otherwise.

\begin{table}
\tablewidth=270pt
\caption{Estimated GEE model for children's mental health data,
from Fitzmaurice, Laird and Ware (\citeyear{FLW04}), page 438}
\label{tabccsgee}
\begin{tabular*}{\tablewidth}{@{\extracolsep{\fill}}ld{2.3}cd{3.2}@{}}
\hline
\textbf{Variable} & \multicolumn{1}{c}{\textbf{Estimate}}
& \multicolumn{1}{c}{\textbf{SE}} & \multicolumn{1}{c@{}}{$\bolds{Z}$} \\
\hline
Intercept & -1.685 & 0.100 & -16.85 \\
Parent informant ($X_1$) & -0.467 & 0.118 & -3.96 \\
Single parent status ($X_2$) & 0.611 & 0.108 & 5.68 \\
Fair or poor child health ($X_3$) & 0.146 & 0.135 & 1.08 \\
Informant${}\times{}$child health ($X_1 X_3$) & 0.452 & 0.157 & 2.87 \\
\hline
\end{tabular*}
\end{table}

Table~\ref{tabccsgee} shows the estimated coefficients from
Fitzmaurice, Laird and Ware (\citeyear{FLW04}), page 438. It suggests
that a report of externalizing behavior is more likely if the informant
is a teacher or the parent is single. The significant interaction
implies that the probability is further increased if the informant is a
parent and the child has fair or poor health.

\begin{figure}

\includegraphics{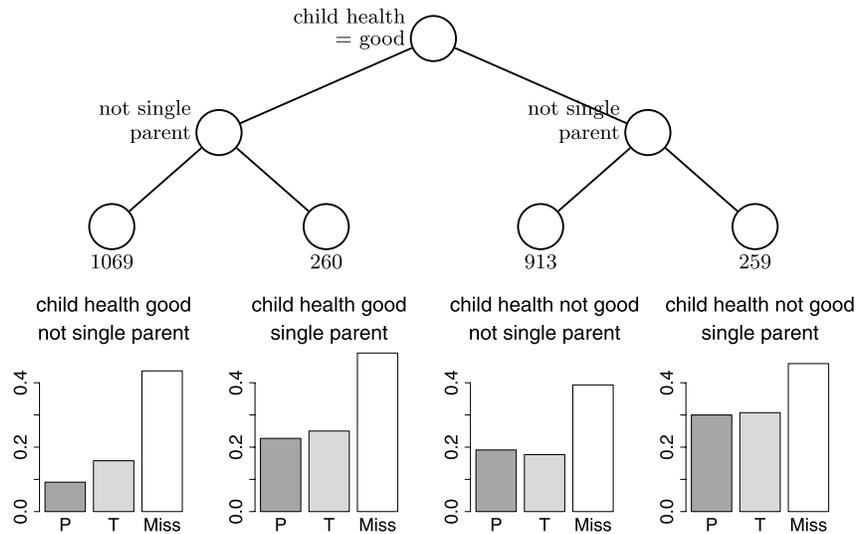}

\caption{Multivariate GUIDE tree model for children's mental health
data. A case goes to the left branch at each intermediate node if
and only if the condition on its left is satisfied. Sample sizes
are given beneath the terminal nodes. The barplots below them give
the proportions of parents (P) and teachers (T) reporting
externalizing behavior and the proportions missing teacher
responses.}
\label{figccsguide}
\end{figure}

The multivariate GUIDE model, using $Z_k = -1$ for missing $Y_k$
values in step~\ref{itemmissy} of Algorithm~\ref{algmult}, is shown
in Figure~\ref{figccsguide}. It splits first on child health and
then on single parent status. (The model using $Z_k = 1$ for missing
$Y_k$ splits first on single parent status and then on child health,
but its set of terminal nodes is the same.) The barplots below the
terminal nodes compare the predicted proportions (means of $Y_{ij}$)
of the parents and teachers who report externalizing behavior and the
proportions of missing teacher responses. The interaction effect in
the GEE model can be explained by the barplots: parent reports of
externalizing behavior are less frequent than teacher reports except
when the child's health is not good and the parent is not single. The
main effect of single parent status is also clear: both parent and
teacher reports are more frequent if the parent is single. Further,
children of single parents are more likely to be missing teacher
%
\begin{figure}[b]

\includegraphics{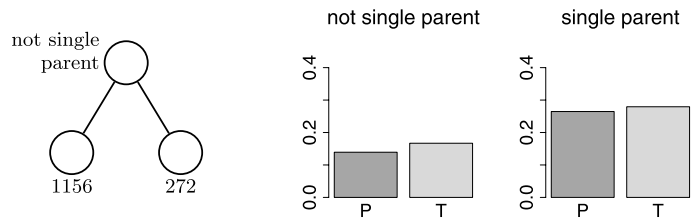}

\caption{MVPART tree model for children's mental health data. A case
goes to the left branch at each intermediate node if and only if
the condition on its left is satisfied. Sample sizes are given
beneath the terminal nodes. The barplots give the proportions of
parents (P) and teachers (T) reporting externalizing behavior and
the proportions missing teacher responses in the terminal
nodes. The model uses only the cases with nonmissing response values.}
\label{figccsmvpart}
\end{figure}
reports. Figure~\ref{figccsmvpart} shows the MVPART tree, which
splits only once, on single parent status. One reason for its brevity
is that it ignores data from the 1073 children that do not have
teacher responses.

\section{Longitudinal data}
\label{seclongitudinal}

Algorithm~\ref{algmult} is directly applicable to longitudinal data as
long as they are observed on a fixed grid and the number of grid points
is small.\vadjust{\goodbreak} Since these conditions may be too restrictive, we show here
how to modify the algorithm for broader applicability. To motivate and
explain the changes, consider a longitudinal study on the hourly wage
of 888 male high school dropouts (246 black, 204 Hispanic, 438 white),
whose observation time points as well as their number (1--13) varied
across individuals. \citeauthor{SW03} [(\citeyear{SW03}),
Section~5.2.1] fit a linear mixed effect (LME) model to the natural
logarithm of hourly wage (\texttt{wage}) to these data. They choose the
transformation partly to overcome the range restriction on hourly wage
and partly to satisfy the linearity assumption. Their model is
%
\begin{eqnarray}
\label{eqdropoutlme} E\log(\mathtt{wage}) & = & \beta_0 +
\beta_1 \mathtt{hgc} + \beta_2 \mathtt{exper} +
\beta_3 \mathtt{black} + \beta_4 \mathtt{hisp}
\nonumber\\
&&{} + \beta_5 \mathtt{exper} \times\mathtt{black} +
\beta_6 \mathtt{exper} \times\mathtt{hisp}
\\
&&{} + b_0 + b_1 \mathtt{exper},
\nonumber
\end{eqnarray}
where \texttt{hgc} is the highest grade completed, \texttt{exper} is
the number of years (to the nearest day, after labor force entry),
\texttt{black}${}={}$1 if a subject is black and 0 otherwise,
%
\begin{table}
\caption{Fixed-effect estimates for linear mixed effect model
(\protect\ref{eqdropoutlme})
fitted to high school dropout data}
\label{tabdropoutlme}
\begin{tabular*}{\tablewidth}{@{\extracolsep{\fill}}ld{2.3}crd{2.2}c@{}}
\hline
& \multicolumn{1}{c}{\textbf{Value}} & \multicolumn{1}{c}{\textbf{Std. error}}
& \multicolumn{1}{c}{\textbf{DF}} &
\multicolumn{1}{c}{$\bolds{t}$\textbf{-value}} &
\multicolumn{1}{c@{}}{$\bolds{p}$\textbf{-value}} \\
\hline
(Intercept) & 1.382& 0.059& 5511& 23.43 & 0.000\\
hgc & 0.038& 0.006& 884& 5.94 & 0.000\\
exper & 0.047& 0.003& 5511& 14.57 & 0.000\\
black & 0.006& 0.025& 884& 0.25 & 0.804\\
hisp & -0.028& 0.027& 884& -1.03 & 0.302\\
exper${}\times{}$black & -0.015& 0.006& 5511& -2.65 & 0.008\\
exper${}\times{}$hisp & 0.009& 0.006& 5511& 1.51 & 0.131\\
\hline
\end{tabular*}
\end{table}
\texttt{hisp}${}={}$1 if a subject is Hispanic and 0 otherwise, and $b_0$
and $b_1$ are subject random effects. The fixed-effect estimates in
Table~\ref{tabdropoutlme} show that \texttt{hgc} and \texttt{exper}
are statistically significant, as is the interaction between
\texttt{exper} and \texttt{black}. The main and interaction effects of
\texttt{hisp} are not significant.

\begin{figure}

\includegraphics{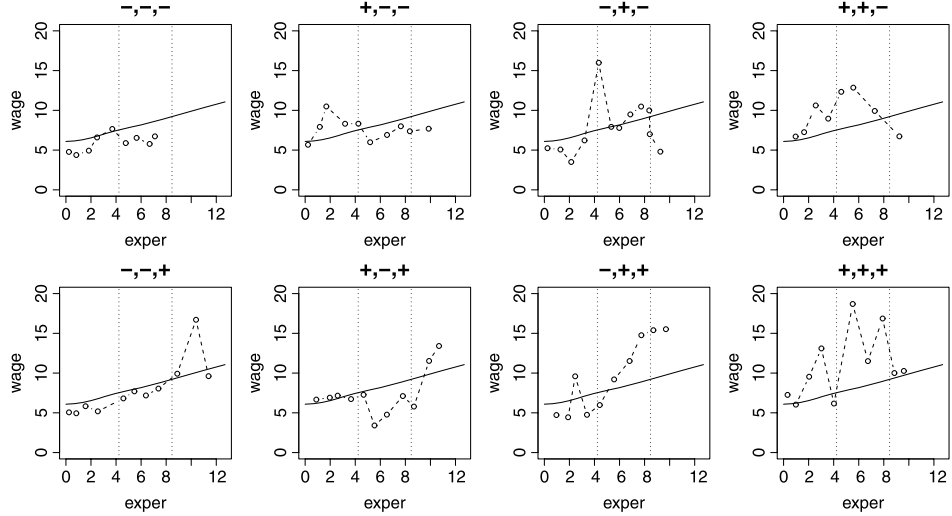}

\caption{Trajectories of eight high school individuals. The solid
curve is the lowess fit to all the subjects. The signs in the plot
titles are the signed values of $(Z_1, Z_2, Z_3)$, where $Z_k = 1$
if the number of observations above the lowess curve is greater
than the number below the curve in the $k$th time interval, and
$Z_k = -1$ otherwise.}
\label{figdropoutsamples}
\end{figure}

Let $Y_{ij}$ denote the response of the $i$th subject at the $j$th
observation time $u_{ij}$. To render Algorithm~\ref{algmult}
applicable to varying numbers and values of $u_{ij}$, we first divide
the range of the $u_{ij}$ values into $d$ disjoint intervals, $U_1,
U_2,\ldots, U_d$, of equal length, where $d$ is user selectable. Then
we replace steps (1) and (2) of the algorithm with these two steps:
\begin{longlist}[(2)]
\item[(1)] At each node, apply the lowess [\citet{lowess}] method to the data
points $(u_{ij}, Y_{ij})$ to estimate the mean of the $Y_{ij}$
values with a smooth curve $S(u)$.
\item[(2)] Define $Z_k = 1$ for subject $i$ if the number of observations
with $Y_{ij} > S(u_{ij})$ is greater than or equal to the number
with $Y_{ij} \leq S(u_{ij})$, for $u_{ij} \in U_k$, $k = 1,2,\ldots, d$. Otherwise, define $Z_k = -1$. (By this definition, $Z_k
= -1$ if there are no observations in $U_k$.)
\end{longlist}

\begin{figure}

\includegraphics{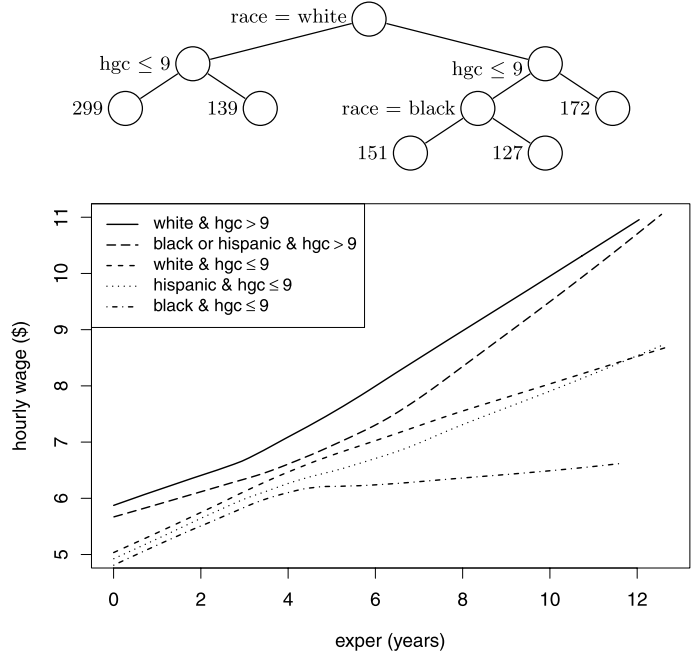}

\caption{Multivariate GUIDE tree for high school dropout data on
top; lowess-smoothed estimates of mean hourly wage by leaf node
on bottom. At an intermediate node, a case goes to the left branch
if and only if the given condition is satisfied; sample sizes are
given beneath the terminal nodes. }
\label{figdropouttree}
\end{figure}

With these changes, we can fit a regression tree model to the wage
data. Since our method is not limited by range restrictions on
$Y_{ij}$ or linearity assumptions, we fit the model to untransformed
hourly wage, using \texttt{hgc} and \texttt{race} as split variables,
\texttt{exper} as the time variable, and $d =
3$. Figure~\ref{figdropoutsamples} shows the lowess curve for the
data at the root node and a sample trajectory for each of the eight
possible values of $(Z_1, Z_2, Z_3)$. Figure~\ref{figdropouttree}
gives the pruned tree, which has five terminal nodes. The first split
is on \texttt{race}; if \texttt{race}${}={}$\texttt{white}, the node is
further split on \texttt{hgc}${}\leq{}$9. Lowess curves for the five
terminal nodes are drawn below the tree. Contrary to the finding in
\citeauthor{SW03} [(\citeyear{SW03}), page 149] that the trajectories of Hispanic and White
subjects cannot be distinguished statistically, we see that Hispanics
tend to have slightly lower hourly wage rates than Whites. In
addition, the slope of the mean trajectory for Blacks with
\texttt{hgc}${}\leq{}$9 appears to decrease after 4 years of experience,
contradicting the exponential trend implied by the logarithmic
transformation of \texttt{wage} in the linear mixed model.

\section{GEE and LME versus GUIDE}
\label{secgee}
A simulation experiment was performed to compare the prediction
accuracies of GEE, GUIDE and LME. Two simulation models are used,
each with five independent predictor variables, $X_1, X_2,\ldots,
X_5$, uniformly distributed on $(-1, 1)$. Longitudinal observations
are drawn at $d$ equally spaced time points, $u = 1, 2,\ldots, d$,
with $d=10$. The models are
%
\begin{equation}\label{eqmodel1}
Y_{u} = 1+X_{1}+X_{2}+2 X_{1}
X_{2}+ 0.5u+b_0 +b_1 u +
\varepsilon_{u}
\end{equation}
and
%
\begin{equation}\label{eqmodel2}
Y_{u} = 2.5I(X_{1} \leq0) + 0.5u +b_0
+b_1 u +\varepsilon_{u},
\end{equation}
where $b_0 \sim N(0,0.5^2)$ and $b_1 \sim N(0,0.25^2)$ are random
effects, $\varepsilon_{u}$ is standard normal, and all are mutually
independent. The fitted model in both cases is
\[
Y_{u} = \beta_0+\beta_1 X_{1}+
\beta_2 X_{2}+\cdots+\beta_5 X_{5}+
\beta_6 u +b_0+b_1 u +
\varepsilon_{u}
\]
and the parameters, $\beta_0,
\beta_1,\ldots, \beta_6$, are estimated using the R packages
\textit{lme4} [\citet{lme4}] and \textit{geepack} [\citet{geepack}] for LME
and GEE, respectively, with GEE employing a compound symmetry
correlation structure. Model (\ref{eqmodel1}) is almost perfect for
LME and GEE except for the interaction term and
model (\ref{eqmodel2}) is almost perfect for GUIDE except for the
terms linear in $u$.

For each simulation trial, a training set of two hundred longitudinal
data series are generated from the appropriate simulation model.
Estimates $\hat{f}(u, x_1, x_2,\ldots, x_5)$ of the conditional mean
$E(y_u| x_1, x_2,\ldots, x_5)$ are obtained for each method on a
uniform grid of $m = 6^5 = 7776$ points $(x_{i1},
x_{i2},\ldots,
x_{i5}) \in(-1, 1)^5$ and the mean squared error
\[
\mathrm{MSE} = (dm)^{-1} \sum_{i=1}^m
\sum_{u=1}^d \bigl\{\hat{f}(u,
x_{i1}, x_{i2},\ldots, x_{i5}) -
E(y_u | x_{i1}, x_{i2},\ldots, x_{i5})
\bigr\}^2
\]
recorded.
Table~\ref{tabgee} shows the average values of the MSE and their
estimated standard errors from 200 simulation trials. There is no
uniformly best method. LME is best in model (\ref{eqmodel1}) and
GUIDE is best in model (\ref{eqmodel2}). Because it makes fewer
assumptions, GEE has a slightly higher MSE than LME in both models.

\begin{table}
\tablewidth=330pt
\caption{Estimated mean squared errors for LME, GEE and GUIDE with
standard errors}
\label{tabgee}
\begin{tabular*}{\tablewidth}{@{\extracolsep{\fill}}lccc@{}}
\hline
& \textbf{LME} & \textbf{GEE} & \textbf{GUIDE} \\
\hline
Model (\ref{eqmodel1}) & $1.00 \pm0.01$ & $1.12 \pm0.01$ & $1.27
\pm0.03$ \\
Model (\ref{eqmodel2}) & $0.49 \pm0.01$ & $0.60\pm0.01$ & $0.12 \pm
0.01$\\
\hline
\end{tabular*}
\end{table}

\section{Time-varying covariates and multiple series}
\label{secparallel}
\begin{figure}

\includegraphics{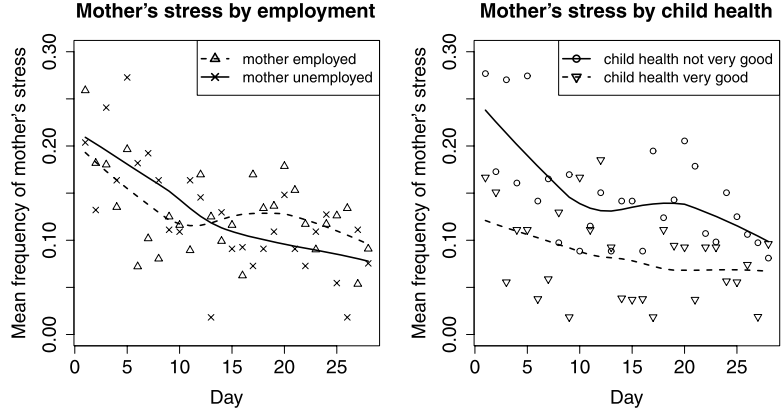}

\caption{Average and smoothed frequencies of mother's stress by
employment and child health.}
\label{figmscmstress}
\end{figure}

Our approach requires all predictor variables to be fixed with respect
to time. An example where there is a time-varying covariate is the
Mothers' Stress and Children's Morbidity study reported in
\citet{am86} and analyzed in \citet{DHLZ02}, Chapter 12.
In this study,
the daily presence or absence of maternal stress and child illness in
167 mother-child pairs was observed over a four-week period. The
children ranged in age from 18 months to 5 years. Time-independent
variables, measured at the start of the study, are mother's marital
and employment status (both binary), education level and health (both
ordinal with 5 categories), child's race and sex (both binary),
child's health (ordinal with 5 categories) and household size (3 or
fewer vs. more than 3 people). \citet{DHLZ02} use GEE logistic
regression models to answer the following questions:
\begin{longlist}[(2)]
\item[(1)] Is there an association between mother's employment and child
illness?
\item[(2)] Is there an association between mother's employment and stress?
\item[(3)] Does mother's stress cause child illness or vice versa?
\end{longlist}
For predicting child illness, their GEE model shows that day (since
enrollment), mother's marital status, child's health and race, and
household size are statistically significant, but mother's employment
is not. Our method gives a trivial tree with no splits after
pruning, suggesting that no variable other than day has predictive
power. For predicting mother's stress, their GEE model finds that
day, mother's health, marital status and education, child's health,
household size and the interaction between day and employment are
significant. Our pruned tree has two terminal nodes, separating
children that have very good health from those that do not.
Figure~\ref{figmscmstress} shows plots of the observed and
lowess-smoothed mean frequencies of mother's stress, grouped by
mother's employment status (left) and by child health as found by our
tree model (right). The curves\vadjust{\goodbreak} defined by employment cross over,
lending support to the significance of the day-employment interaction
effect found in the GEE model. The large separation between the two
curves defined by child's health, on the other hand, indicates a large
main effect.

\begin{figure}

\includegraphics{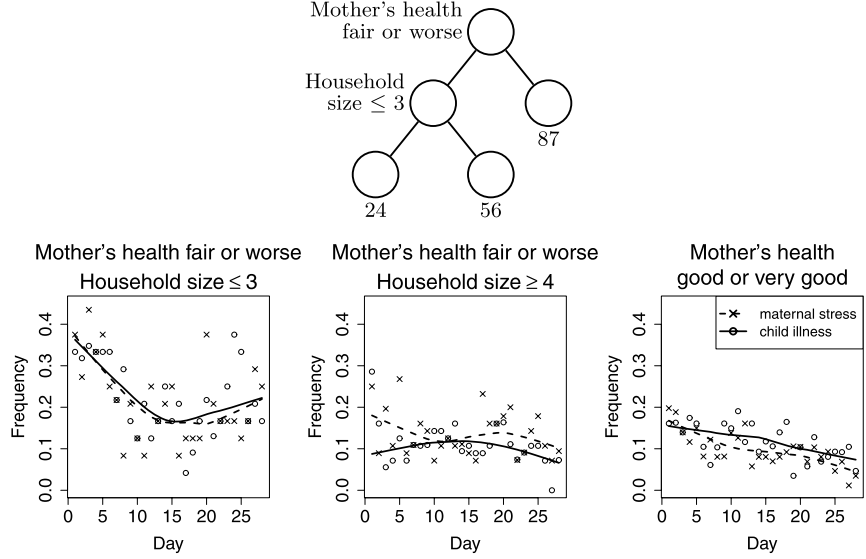}

\caption{Multivariate GUIDE model for simultaneously predicting
maternal stress and child health. A case goes to the left branch
at each intermediate node if and only if the condition on its left
is satisfied. The number beneath each terminal node is the sample
size. The plots below the tree show the observed and smoothed
daily mean frequencies of mother's stress and child's illness.}
\label{figmscmtree}\vspace*{-3pt}
\end{figure}

On the third question of whether mother's stress causes child's
illness or vice versa, \citet{DHLZ02} find, by fitting GEE models with
lagged values of stress and illness as additional predictors, that the
answer can be both. They conclude that there is evidence of feedback,
where a covariate both influences and is influenced by a
response. Instead of trying to determine which is the cause and which
is the effect, we fit a regression tree model that simultaneously
predicts mother's stress and child's illness by concatenating the two
series into one long series with 56 observations. Choosing $d=8$ (four
intervals each for stress and illness), we obtain the results in
Figure~\ref{figmscmtree}, which shows that mother's health and
household size are the most important predictors. The plots below the
%
tree confirm that mother's stress (dashed curves) and child's illness
(solid curves) vary together. More interesting is that the two
responses do not decrease monotonically with time. In particular, when
mother's health is fair or worse and household size is three or less,
the frequencies of mother's stress and child's illness tend to
decrease together in the first half and increase together in the
second half of the study period. This behavior is ruled out by the
GEE model of \citet{DHLZ02}. We are thus reminded that the statistical
significance of the terms in a parametric model always depends on the
model being correctly specified. If the specification is correct, the
parametric approach will often possess greater sensitivity; otherwise
important features of the data may be undetected.\looseness=-1

\section{Asymptotic consistency}
\label{sectheory}
We give some conditions for asymptotic consistency of the regression
function estimates, as the training sample size increases, for
multiresponse and longitudinal data models.\vadjust{\goodbreak} The conditions generalize
those for univariate responses in Chaudhuri et~al.
(\citeyear{chly94,clly95}),
\citet{quart} and \citet{iie07}. We assume that there is a true
regression function $g(\mathbf{x},u)$, where $\mathbf{x}$ is a vector
of predictor variable values in a compact set, $u$ is the observation
time in a compact set $U$, and $\sup_{u,\mathbf{x}} |g(\mathbf{x},u)|
< \infty$. The training data consist of vectors $(y_{ij},
\mathbf{x}_i, u_{ij})$, $i=1,\ldots,M$ and $j=1,\ldots,m_i$, where
$y_{ij}$ is the observed response of subject $i$ at time $u_{ij} \in
U$, $\mathbf{x}_i$ is the corresponding $\mathbf{x}$ value, and
$y_{ij}=g(\mathbf{x}_i,u_{ij})+\varepsilon_{ij}$. The $\varepsilon_{ij}$'s
are assumed to have zero mean, constant (finite) variance and to be
independent of $u_{ij}$ for all $i$ and $j$. This setup applies to the
multiresponse model as well, because it can be treated as a
longitudinal model with fixed time points. Let $N=\sum_{i=1}^M m_i$
denote the total number of data points and let $T_N$ denote the
collection of terminal nodes of a regression tree obtained by
partitioning the data by its $\mathbf{x}$ values. Given
$(\mathbf{x}^*, u^*)$, let $t^*$ denote the terminal node containing
$\mathbf{x}^*$.\vspace*{-3pt}

\subsection{Multiresponse and longitudinal data with fixed time
points}
Assume that $U$ is a finite set. Let $\delta(T_N) = \min_{t\in T_N, u
\in U}|\{(i,j)\dvtx  \mathbf{x}_i \in t, u_{ij} = u \}|$ denote the
smallest number of data points per time point across all terminal
nodes. Define $I_N^* = \{(i,j)\dvtx  \mathbf{x}_i\in t^*, u_{ij}=u^*\}$
and let $k_N$ denote the number of elements in $I_N^*$. Assume further that
the following conditions hold:
\begin{longlist}[(A3)]
\item[(A1)] The $\varepsilon_{ij}$ are mutually independent for all $i$
and $j$.\vadjust{\goodbreak}
\item[(A2)] $\delta(T_N) \stackrel{P}{\rightarrow} \infty$ as $N
\rightarrow\infty$.
\item[(A3)] For each $u \in U$, $\sup_{t\in
T_N}\sup_{\mathbf{x}_1,\mathbf{x}_2\in t} |g(\mathbf{x}_1,u)-
g(\mathbf{x}_2,u)| \stackrel{P}{\rightarrow} 0$ as \mbox{$N
\rightarrow\infty$}.
\end{longlist}
Condition (A2) ensures that there are sufficient observations in each
terminal node for consistent estimation. Condition (A3) requires the
function to be sufficiently smooth; it implies that for each $u \in
U$, $g(\mathbf{x},u)$ is uniformly continuous w.r.t. $\mathbf{x}$ in
each $t \in T_N$. In other words, (A3) assumes that the partitioning
algorithm is capable of choosing the right splits so that within each
node, the mean response curves are close to each other.

The regression estimate of $g(\mathbf{x}^*,u^*)$ is
\begin{eqnarray*}
\hat{g}\bigl(\mathbf{x}^*,u^*\bigr) & = & k_N^{-1}\sum
_{(i,j) \in I_N^*} y_{ij}
\\
& = & k_N^{-1}\sum_{(i,j) \in I_N^*} \bigl
\{g(\mathbf{x}_i, u_{ij}) + \varepsilon_{ij}\bigr\}
\\
& = & k_N^{-1}\sum_{(i,j) \in I_N^*} \bigl
\{g\bigl(\mathbf{x}_i, u^*\bigr) + \varepsilon_{ij}\bigr\}
\end{eqnarray*}
by definition of $I_N^*$. Therefore,
\[
\bigl|\hat{g}\bigl(\mathbf{x}^*,u^*\bigr)-g\bigl(\mathbf{x}^*,u^*\bigr)\bigr| \leq
k_N^{-1} \biggl|\sum_{(i,j) \in I_N^*} \bigl\{g
\bigl(\mathbf {x}_i,u^*\bigr)-g\bigl(\mathbf{x}^*,u^*\bigr)\bigr\} \biggr|
+ k_N^{-1} \biggl|\sum_{(i,j) \in I_N^*}
\varepsilon_{ij} \biggr|.
\]
Condition (A3)
implies that the first term on the right-hand side of the inequality
converges to zero in probability. Condition (A2) implies that $k_N
\stackrel{P}{\rightarrow} \infty$, which together with the
independence and constant variance assumptions on $\varepsilon_i$ imply
that the second term converges to zero as well. Therefore,
$\hat{g}(\mathbf{x}^*,u^*) \stackrel{P}{\rightarrow}
g(\mathbf{x}^*,u^*)$ as $N \rightarrow\infty$ at every
$(\mathbf{x}^*, u^*)$.

\subsection{Longitudinal data with random time points}
Suppose now that $U$ is a compact interval and that the $u_{ij}$'s are
random. Let $K(u) \geq0$ be a kernel function with bandwidth
$h_N$. The estimate of $g(\mathbf{x},u)$ at $(\mathbf{x}^*, u^*)$ is
\[
\hat{g}\bigl(\mathbf{x}^*,u^*\bigr) = \frac{\sum_{\mathbf{x}_i \in
t^*}\sum_{j=1}^{m_i}K\{h_N^{-1} (u_{ij}-u^*)\}y_{ij}} {
\sum_{\mathbf{x}_i \in t^*}\sum_{j=1}^{m_i}K\{h_N^{-1}
(u_{ij}-u^*)\}}.
\]
Let $n_N$ denote the smallest number of data
points in the terminal nodes of the tree. Assume that the following
conditions hold:
\begin{enumerate}[(B4)]
\item[(B1)] The $u_{ij}$ values are independent and identically
distributed and their density function $f(u)$ is positive everywhere
and does not depend on the $\mathbf{x}_i$ values, for all $i$ and
$j$.
\item[(B2)] $\sup_{t\in T_N}\sup\{|g(\mathbf{x}_1,u)-
g(\mathbf{x}_2,u)|\dvtx  u \in U,   \mathbf{x}_1,\mathbf{x}_2\in t \}
\stackrel{P}{\rightarrow} 0$ as $N \rightarrow\infty$.\vadjust{\goodbreak}
\item[(B3)] The density function of $u_{ij}$ is positive everywhere in
$U$ and:
\begin{enumerate}[(iii)]
\item[(i)] $\int|K(u)|  \,du < \infty$,
\item[(ii)] $\lim_{|u| \rightarrow\infty} uK(u) = 0$,
\item[(iii)] $n_N \stackrel{P}{\rightarrow} \infty$, $h_N
\stackrel{P}{\rightarrow} 0$ and $n_N h_N
\stackrel{P}{\rightarrow} \infty$ as $N \rightarrow\infty$.
\end{enumerate}
\item[(B4)] The error vectors
$\varepsilon_i=(\varepsilon_{i1},\ldots,\varepsilon_{im_i})'$ are independent
between subjects. For each $i$, $\varepsilon_i$ has a covariance matrix
with elements $\sigma_{ijk}$ such that $\sigma_{ijk} = \sigma^2$ for
$j = k$ and $\max_i m_i^{-1} \sum_{j\neq k} \sigma_{ijk} \leq A$ for
some positive constant $A$.
\end{enumerate}
Condition (B1) ensures that the value of $u^*$ is not constrained by the
value of~$\mathbf{x}^*$. Condition (B2) is a stronger version of (A3) and
condition (B3) is a standard requirement for consistency of kernel
estimates. Condition (B4) ensures that the correlations between the
random errors are small.

Write
\begin{eqnarray*}
&&\hat{g}\bigl(\mathbf{x}^*,u^*\bigr)-g\bigl(\mathbf{x}^*,u^*\bigr)
\\
&&\qquad= \frac{
\sum_{\mathbf{x}_i \in t^*}\sum_{j=1}^{m_i}K\{h_N^{-1}(u_{ij}-u^*)\}
\{y_{ij}-g(\mathbf{x}^*,u^*)\}} {
\sum_{\mathbf{x}_i \in t^*}\sum_{j=1}^{m_i}K\{h_N^{-1}(u_{ij}-u^*)\}
}
\\
&&\qquad= \frac{\sum_{\mathbf{x}_i \in t^*}\sum_{j=1}^{m_i}K\{
h_N^{-1}(u_{ij}-u^*)\}
\{g(\mathbf{x}_i,u_{ij}) + \varepsilon_{ij}-g(\mathbf{x}^*,u^*)\}} {
\sum_{\mathbf{x}_i \in t^*}\sum_{j=1}^{m_i}K\{h_N^{-1}(u_{ij}-u^*)\}
}
\\
&&\qquad = \frac{\sum_{\mathbf{x}_i \in t^*}\sum_{j=1}^{m_i}
K\{h_N^{-1}(u_{ij}-u^*)\}\{g(\mathbf{x}^*,u_{ij})-g(\mathbf
{x}^*,u^*)\}
}{\sum_{\mathbf{x}_i \in t^*}\sum_{j=1}^{m_i}K\{
h_N^{-1}(u_{ij}-u^*)\}}
\\
&&\qquad\quad{} + \frac{\sum_{\mathbf{x}_i \in t^*}\sum_{j=1}^{m_i}K\{
h_N^{-1}(u_{ij}-u^*)\}
\{g(\mathbf{x}_i,u_{ij})-g(\mathbf{x}^*,u_{ij})\}} {
\sum_{\mathbf{x}_i \in t^*}\sum_{j=1}^{m_i}K\{h_N^{-1}(u_{ij}-u^*)\}
}
\\
&&\qquad\quad{} + \frac{ \sum_{\mathbf{x}_i \in t^*}\sum_{j=1}^{m_i}K\{
h_N^{-1}(u_{ij}-u^*)\}
\varepsilon_{ij}}{\sum_{\mathbf{x}_i \in t^*}\sum_{j=1}^{m_i}K\{
h_N^{-1}(u_{ij}-u^*)\}}
\\
&&\qquad= J_1+J_2+J_3 \qquad\mbox{(say).}
\end{eqnarray*}
Define the local polynomial estimator (which depends on $u_{ij}$'s but
not on the values of $\mathbf{x}_1,\ldots, \mathbf{x}_M$)
\[
\bar{g}\bigl(\mathbf{x}^*,u^*\bigr) = \frac{\sum_{\mathbf{x}_i \in t^*}\sum_{j=1}^{m_i}K\{h_N^{-1}
(u_{ij}-u^*)\} g(\mathbf{x}^*,u_{ij})} {\sum_{\mathbf{x}_i \in
t^*}\sum_{j=1}^{m_i}K\{h_N^{-1} (u_{ij}-u^*)\}}.
\]
Then $J_1 =
\bar{g}(\mathbf{x}^*,u^*) - g(\mathbf{x}^*,u^*)
\stackrel{P}{\rightarrow} 0$ by condition (B3) [\citet{hardle}, page~29] and
$J_2 \stackrel{P}{\rightarrow} 0$ by condition (B2).

Note that\vspace*{1pt} $(Nh_N)^{-1} \sum_{\mathbf{x}_i \in
t^*}\sum_{j=1}^{m_i}K\{h_N^{-1}(u_{ij}-u^*)\}
\stackrel{P}{\rightarrow} f(u^*)\int K(z)  \,dz$,\break where $f(u)$ is the
density function of the $u_{ij}$. Conditions (B1) and (B4) imply
\begin{eqnarray*}
&&E \Biggl[ (Nh_N)^{-1} \sum_{\mathbf{x}_i \in t^*}
\sum_{j=1}^{m_i}K\bigl\{h_N^{-1}
\bigl(u_{ij}-u^*\bigr)\bigr\} \varepsilon_{ij}
\Biggr]^2
\\
&&\qquad= (Nh_N)^{-2} \sum_{\mathbf{x}_i \in t^*}
\sigma^2 m_i E\bigl[K^2\bigl
\{h_N^{-1}\bigl(u_{i1}-u^*\bigr)\bigr\}\bigr]
\\
&&\qquad\quad{} + (Nh_N)^{-2} \sum
_{\mathbf{x}_i \in t^*} \bigl[EK\bigl\{h_N^{-1}
\bigl(u_{i1}-u^*\bigr)\bigr\}\bigr]^2 \sum
_{j\neq k} \sigma_{ijk}
\\
&&\qquad\leq \sigma^2 N^{-1} h_N^{-2} E
\bigl[K^2\bigl\{h_N^{-1}\bigl(u_{i1}-u^*
\bigr)\bigr\}\bigr]
\\
&&\qquad\quad{} + A(Nh_N)^{-2}\bigl[EK\bigl
\{h_N^{-1}\bigl(u_{i1}-u^*\bigr)\bigr\}
\bigr]^2\sum_{\mathbf
{x}_i \in t^*}m_i
\\
&&\qquad= \sigma^2 (Nh_N)^{-1} f\bigl(u^*\bigr)\int
K^2(z) \,dz + AN^{-1} \biggl\{\int K(z) \,dz \biggr
\}^2 +o(1)
\\
&&\qquad \rightarrow 0.
\end{eqnarray*}
It follows that $J_3 \stackrel{P}{\rightarrow} 0$ and, hence,
$\hat{g}(\mathbf{x}^*,u^*) \stackrel{P}{\rightarrow}
g(\mathbf{x}^*,u^*)$ as $N \rightarrow\infty$.

\section{Concluding remarks}
\label{secconclusion}
Previous algorithms for fitting regression trees to multiresponse and
longitudinal data typically follow the CART approach, with various
likelihood-based node impurity functions. Although straightforward,
this strategy has two disadvantages: the algorithms inherit the
variable selection biases of CART and are constrained by computational
difficulties due to maximum likelihood and covariance estimation at
every node of the tree.

To avoid these problems, we have introduced an algorithm based on the
univariate GUIDE method that does not have selection bias and does not
require maximization of likelihoods or estimation of covariance
matrices. Unbiasedness is obtained by selecting the split variable
with contingency table chi-squared tests, where the columns of each
table are defined by the patterns of the data trajectories relative to
the mean trajectory and the rows are defined by the values of a
predictor variable. The mean trajectory is obtained by applying a
nonparametric smoother to the data in the node. For split set
selection and for tree pruning, the node impurity is defined as the
total, over the number of response variables, of the sum of
(optionally normalized) squared errors for each response variable.
Correlations among longitudinal response values are implicitly
accounted for by the smoothing and the residual trajectory patterns.

Because no assumptions are made about the structure of the model in
each node, it is quite possible that our method is less powerful than
other tree methods\vadjust{\goodbreak} in situations where the assumptions required by the
latter are satisfied. (These assumptions, such as autoregressive
models, are hard to justify because they need to be satisfied within
random partitions of the data.) What we lose in sensitivity, though,
we expect to gain in robustness. Besides, the simplicity of our
smoothing and means-based approach lends itself more easily to
asymptotic analysis. Further, as is evident from the longitudinal data
examples, plots of the smoothed mean trajectories in the terminal
nodes provide a visual summary of the data that is more realistic than
the necessarily more stylized summaries of parametric or
semi-parametric models.\looseness=-1

Our approach should not be regarded, however, as a substitute for
parametric and semi-parametric methods such as GEE and LME for
longitudinal data. Because the latter methods assume a parametric
model for the mean response function, they permit parametric
statistical inference, such as significance tests and confidence
intervals, to be performed. No such inference is possible for
regression tree models, as there are no model parameters in the
traditional sense. Regression tree models are simply approximations
to the unknown response functions, whatever they may be, and are meant
for descriptive and prediction purposes. Although GEE and LME models
can be used for prediction too, their constructions are based on
significance tests, unlike tree models which are focused on prediction
error. In applications where the sample size and number of predictor
variables are small and the model is correctly specified, GEE and LME
will always be more powerful than tree methods, due to the extra
information provided by the parametric model. But if the sample size
or the number of predictor variables is large, it can be challenging
to select the right parametric model. It is in such situations that a
regression tree model can be quite useful because it provides a
relatively simple and interpretable description of the data. The
fitted tree model can serve a variable selection purpose as well, by
identifying a subset of predictor variables for subsequent parametric
modeling, if desired.

The proposed method is implemented in the GUIDE software which can be
obtained from
\href{https://www.stat.wisc.edu/\textasciitilde loh/guide.html}{www.stat.wisc.edu/\textasciitilde loh/guide.html}.

\section*{Acknowledgments}

We are grateful to Editor Susan Paddock, an anonymous Associate Editor
and two referees for comments and suggestions that led to improvements
in the article. CART\circledR\ is a registered trademark of
California Statistical Software, Inc.



\printaddresses

\end{document}